%% file: root.tex
\documentclass[letterpaper, 10 pt, conference]{ieeeconf}  

\IEEEoverridecommandlockouts    
\usepackage{hhline}
\usepackage{soul}
\usepackage{xcolor}
\usepackage{cite}
\usepackage{import}
\usepackage{amsmath}
\usepackage{graphicx}
\usepackage{amsfonts} 
\usepackage{bm}
\usepackage[capitalise]{cleveref}

\title{\LARGE \bf
Whole-body MPC for highly redundant legged manipulators: experimental evaluation with a 37 DoF dual-arm quadruped
}

\author{Ioannis Dadiotis, Arturo Laurenzi and Nikos Tsagarakis
\thanks{
This work was supported from the European Union’s Horizon Europe Framework Programme under grant agreement No. 101070596 (euROBIN).
}
\thanks{All authors are with Humanoid and Human-Centered Mechatronics Research Line, Italian Institute of Technology, Genoa 16163, Italy {(emails: \tt\small name.surname@iit.it)}. Ioannis Dadiotis is also with
the Department of Informatics, Bioengineering, Robotics and Systems Engineering, University of Genoa, Genoa 16145, Italy.}
}

\usepackage{tikz}
\usepackage{textcomp}
\usepackage{lipsum}

\usepackage{multirow}   

\newcommand\copyrighttext{%
  \footnotesize \textcopyright 2023 IEEE. Personal use of this material is permitted.
  Permission from IEEE must be obtained for all other uses, in any current or future
  media, including reprinting/republishing this material for advertising or promotional
  purposes, creating new collective works, for resale or redistribution to servers or
  lists, or reuse of any copyrighted component of this work in other works.}
\newcommand\copyrightnotice{%
\begin{tikzpicture}[remember picture,overlay]
\node[anchor=south,yshift=10pt] at (current page.south) {\parbox{\dimexpr\textwidth-\fboxsep-\fboxrule\relax}{\copyrighttext}};
\end{tikzpicture}%
}

\begin{document}
\bstctlcite{IEEEexample:BSTcontrol}

\maketitle
\copyrightnotice

\thispagestyle{empty}
\pagestyle{empty}

\begin{abstract}
Recent progress in legged locomotion has rendered quadruped manipulators a promising solution for performing tasks that require both mobility and manipulation (\emph{loco-manipulation}). In the real world, task specifications and/or environment constraints may require the quadruped manipulator to be equipped with \emph{high redundancy} as well as \emph{whole-body} motion coordination capabilities. This work presents an experimental evaluation of a whole-body Model Predictive Control (MPC) framework achieving real-time performance on a dual-arm quadruped platform consisting of 37 actuated joints. To the best of our knowledge this is the legged manipulator with the highest number of joints to be controlled with real-time whole-body MPC so far. The computational efficiency of the MPC while considering the full robot kinematics and the centroidal dynamics model builds upon an open-source DDP-variant solver and a state-of-the-art optimal control problem formulation. Differently from previous works on quadruped manipulators, the MPC is directly interfaced with the low-level joint impedance controllers without the need of designing an instantaneous whole-body controller. The feasibility on the real hardware is showcased using the CENTAURO platform for the challenging task of picking a heavy object from the ground. Dynamic stepping (trotting) is also showcased for first time with this robot. The results highlight the potential of replanning with whole-body information in a predictive control loop.
\end{abstract}

\input{introduction}
\input{Preliminaries}

\input{Mpc}

\input{FromMPCToMotorControl}

\input{Results}

\input{Conclusion}



\input{references}

\end{document}

%% file: introduction.tex
\section{Introduction}
\begin{figure}
  \centering
  \graphicspath{{figures/}}
  \includegraphics[width=0.85\columnwidth]{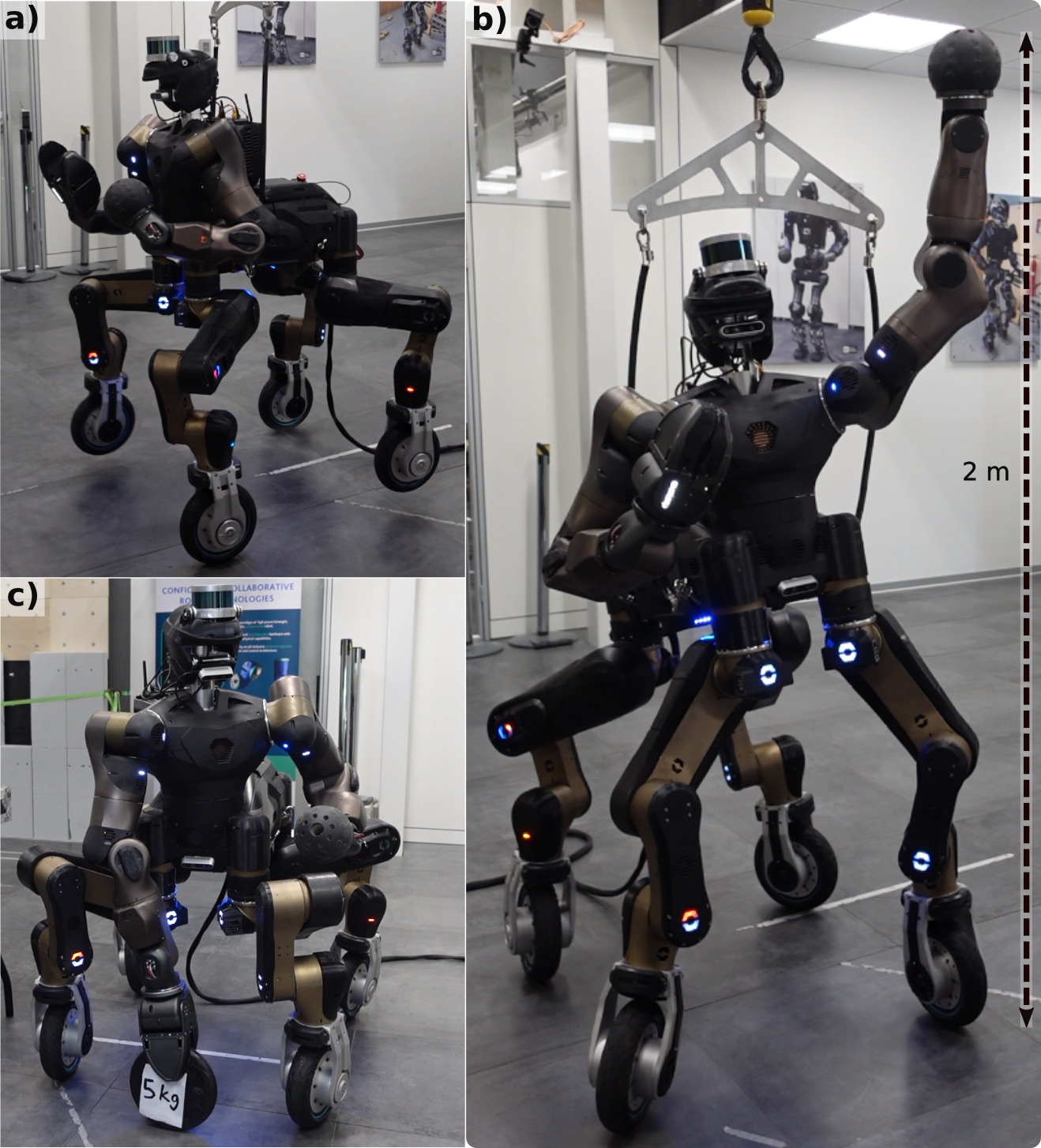}
  \caption{The dual-arm quadruped CENTAURO robot controlled with the proposed whole-body MPC for a variety of motions: a) trotting b) 5 kg object picking from the ground and c) large height reaching. Corresponding video: https://youtu.be/8XIAtw4201o}
  \label{Fig:robot_photo}
\end{figure}
%
Model Predictive Control (MPC) has proved to be a powerful strategy for addressing the motion planning and control problem of quadrupeds, which have now achieved a basic level of autonomy in real world \cite{tranzatto2022cerberus}. The underlying reason of deploying MPC is the intuitiveness to formulate the problem as a constrained model-based optimal control problem (OCP) that is continuously solved over a receding finite time horizon. Optimizing over the time horizon enables anticipating future expected events (e.g. make/break of contacts) and continuous replanning compensates for model errors and environment changes, while satisfying the imposed constraints. Recently there has been an increasing interest in extending these capabilities to quadruped manipulators towards enabling the execution of complex \emph{loco-manipulation} tasks. Legged manipulators, in general, promise the real-world sensorimotor skills needed for interaction not only between the robot legs and the terrain to be traversed (\emph{legged locomotion}) but between one or more mounted robotic arms and its surrounding objects (\emph{manipulation}), as well.

Unfortunately, the promising applicability when augmenting a quadruped with one or more robotic arms comes at the cost of increased complexity, since quadruped manipulators consist of multiple kinematic chains and Degrees-of-Freedom (DoF). This increased system dimension combined with the non-linear, hybrid and floating-base dynamics render real-time MPC a hard problem, which highly depends on the MPC design specifications, e.g. the robot model and the OCP solver. This difficulty has motivated a number of approaches which plan and/or control the motion for locomotion (legged body) and manipulation (armed body) separately, therefore splitting the problem into two tractable ones \cite{bellicoso2019alma, bd_dynamic_manipulation, murphy2012high, yuntao2022, parosi2023kinematically}. In some cases the effect of manipulation planning has been considered during locomotion control but not vice versa \cite{yuntao2022}. These approaches are considerably limited when it comes to real-world tasks, which may require coordination of the robot's whole-body for leveraging the available redundancy, e.g. interacting with objects close to the limit of the robot's workspace while satisfying multiple constraints. In fact, humans' advanced mobile manipulation skills are based on whole-body coordination even for simple tasks like opening a door or picking up an object from the ground.

To that end it is easily understood that a MPC capable for generating \emph{whole-body} motion behaviors should consider the full robot kinematics as well as a dynamics model with as much fidelity as required by the targeted task (ideally full dynamics). In this direction, a number of works on bipeds \cite{mpc_hrp2, dantec2021, dantec2022first}, quadrupeds \cite{neunert2018whole, mastalli_id_mpc, assirelli:hal-03778738, grandia2019feedback} as well as other mobile manipulators \cite{minniti2019whole} have achieved real-time performance. For what concerns quadruped manipulators the work of \cite{sleiman2021unified} (applied, also, to \cite{cheng2022haptic, arcari2023bayesian}) has achieved similar results, considering full kinematics and the centroidal dynamics, for a quadruped with an up to 6 DoF robotic arm\footnote{Most of the mentioned works rely on recent advances on rigid multi-body dynamics software \cite{carpentier2019pinocchio} as well as numerical Differential Dynamic Programming (DDP) based solvers \cite{mastalli_crocoddyl}, \cite{OCS2}.}. Despite this progress, whole-body MPC has not been yet evaluated for quadruped manipulators that go beyong the usual 3 DoF leg structure and a single 6 DoF mounted arm, which can be only partially explained by the limited current availability of this kind of platforms. In fact, there exist highly redundant quadrupeds (bi-manual and/or with more complex leg kinematics) but they are mainly planned and controlled through simplified model-based techniques \cite{dadiotis2022trajectory} or teleoperation schemes \cite{klamt2020remote, gitai_video} that rely on instantaneous control (i.e. there is no prediction). As a result, it is not clear to what extent, in terms of real-time performance, whole-body MPC can be applied to such systems and the advantages of this method have not been deployed for a variety of applications that require higher redundancy. Such tasks can be bi-manual loco-manipulation ones (where at least two arms are needed) or heavy physical interaction tasks (e.g. where higher-DoF legged articulation can establish suitable support contacts \cite{polverini2020multi}). This paper consists a first step towards this direction.

\begin{figure*}[t]
  \centering
  \graphicspath{{figures/}}
  \includegraphics[width=0.9\textwidth]{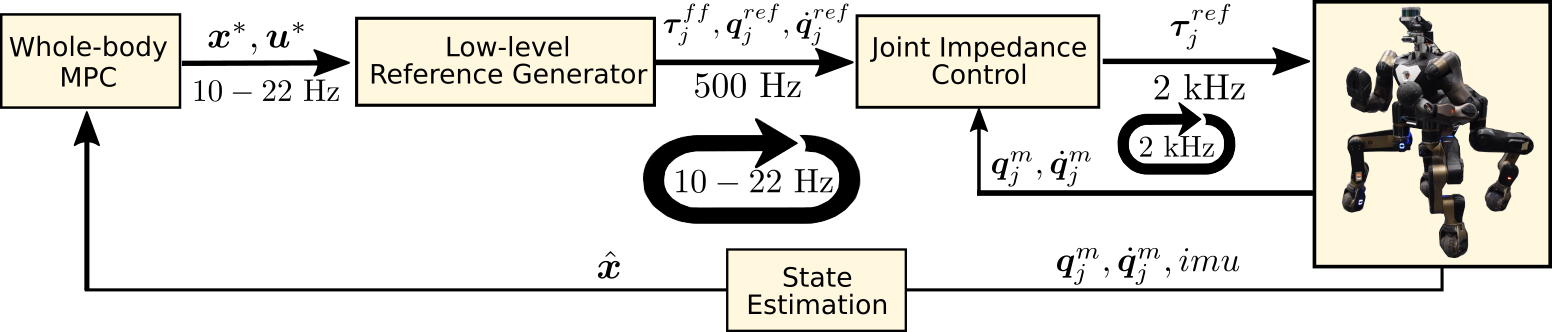}
  \caption{The motion planning and control framework for the dual-arm quadruped CENTAURO platform. The whole-body MPC, considering the centroidal dynamics, computes optimal state and control input trajectories at 10-22 Hz which are then mapped to references for the joint impedance controllers through an inverse dynamics-based low-level reference generator module. Each MPC iteration is updated with the latest available state estimate.}
  \label{Fig:framework}
\end{figure*}

The main contribution of this work is presenting the experimental results of applying whole-body MPC and achieving real-time performance to a highly redundant dual-arm quadruped. The aim is to bridge the gap between motion control of today's most redundant quadruped manipulators and current MPC capabilities. The adopted approach is inspired from \cite{sleiman2021unified}, however the CENTAURO platform \cite{kashiri2019centauro} used in this work consists, at its current version, of much more actuated joints. In particular, it has 37 actuated DoFs which is more than double than the 18 actuated DoFs quadruped manipulator of \cite{sleiman2021unified, cheng2022haptic, arcari2023bayesian} and much higher than the 22 actuated DoFs considered in state-of-the-art MPC schemes for bipeds \cite{dantec2021, dantec2022first}. Besides the technical details for scaling whole-body MPC to such a high-dimensional system the proposed framework, differently from \cite{sleiman2021unified, cheng2022haptic, dadiotis2022trajectory},  does not rely on a WBC. This is only feasible due to the whole-body information considered in the MPC formulation. The motivation for exploring an approach without a WBC arises from the fact that the latter is a separate controller with different control objective and constraints, thus, does not explicitly satisfy compliance with the optimal MPC policy, as has been argued in \cite{mastalli2022agile}. Finally tuning a WBC consists an additional task requiring high expertise. Summing up the contributions of this paper are:
\begin{itemize}
\item An implementation and experimental evaluation study using a whole-body MPC framework for a highly redundant dual-arm quadruped that considers full kinematics as well as the centroidal dynamics. To the best of our knowledge this is the most redundant quadruped manipulation platform (37 actuated DoFs) that has been ever controlled using real-time whole-body MPC. 
\item  A MPC-based pipeline providing motion plans that are executed without an instantaneous WBC. Shown in \cref{Fig:framework}, the pipeline is characterized by two closed loops: the whole-body MPC loop and the joints' impedance control loop at a lower level. The MPC is interfaced with the low-level joint impedance controllers through a low-level reference generator module.
\end{itemize}

%% file: Preliminaries.tex
\section{Preliminaries}
\label{sec:system}
\subsection{The CENTAURO robot}
\label{sec:centauro}
CENTAURO robot is a human-size dual-arm quadruped entirely designed at the Italian Institute of Technology \cite{kashiri2019centauro}. Its lower body consists of four 6 DoF legs. The first 5 leg joints can be controlled in joint position or impedance control mode while the last one is an actuated wheel, which is controlled in velocity mode. Although in this work no wheeled motion is studied, wheels are considered in the OCP so that the formulation can be easily extended by just imposing different contact constraints. The robot at its current version is equipped with two 6 DoF arms (one DoF less with respect to (wrt) the initial 7 DoF prototype). Finally, the robot has a torso yaw joint that connects the upper body with the base of the platform. The above make up a total 37 actuated DoF platform with weight 118 kg.
\subsection{Robot model}
\label{sec:model}
In this paper, the approach of \cite{sleiman2021unified} is adopted where the full kinematics and the centroidal dynamics \cite{orin2013centroidal} are considered while the use of Single Rigid Body Dynamics (SRBD) model \cite{grandia2019feedback} is, also, evaluated. The target is to demonstrate how whole-body MPC scales to such a highly redundant robot and not a comparison of the two dynamic models (which has been already presented in \cite{sleiman2021unified}). We briefly revisit the modeling approach; the reader is advised to the original work for detailed presentation. The dynamic model considered for CENTAURO robot assumes that the robot has six possible contact points, four at the leg and two at the arm EEs. The Newton-Euler equations can be, therefore, expressed wrt the center of mass (CoM) as shown in \cref{eq:CentroidalDynamics}, where $\bm h_{com} \in \mathbb{R}^{6}$ is the centroidal momentum about a reference frame attached to the robot's center of mass and aligned with the world frame. $\bm r_{com, c_{i}}$ is defined as the position of the contact $c_i$ wrt the center of mass, $\bm f_{c_i}$ is the corresponding contact force vector and $n_c = 6$ the number of EEs. Additionally, centroidal momentum can be expressed wrt to the joint position and velocity vectors, through the centroidal momentum matrix (CMM) $\bm A(\bm q) \in \mathbb{R}^{6\times(6+n_a)}$ with $n_a=37$, as shown in \cref{eq:BaseDerivative}. The CMM $\bm A(\bm q) = [\bm A_{b}(\bm q) \ \ \bm A_j(\bm q)]$ consists of the base part $\bm A_{b}(\bm q)$ and the actuated joints part $\bm A_{j}(\bm q)$. Full kinematics are considered since the base pose $\bm q_b \in \mathbb{R}^6$ and actuated joint positions vector $\bm q_j \in \mathbb{R}^{n_a}$ emerge in \cref{eq:BaseDerivative}.
\begin{align} 
    \dot{\bm h}_{com} &= 
    \begin{bmatrix} m \bm g + \sum\limits_{i = 1}^{n_c} \bm f_{c_i}\\
    \sum\limits_{i = 1}^{n_c} \bm r_{com, c_i} \times \bm f_{c_i}
    \end{bmatrix}
    \label{eq:CentroidalDynamics}\\
    \dot{\bm q}_b &= \bm A^{-1}_b \left(\bm h_{com} - \bm A_j \dot{\bm q}_j \right)
    \label{eq:BaseDerivative}
\end{align}
\cref{eq:CentroidalDynamics} and \cref{eq:BaseDerivative} describe the robot model that is considered in the MPC. By selecting the state vector of the model ${\bm x = (\bm h_{com}, \ \bm q_b, \ \bm q_j) \in \mathbb{R}^{12 + n_a}}$ and control input vector ${\bm u = (\bm f_{c_1}, \ ..., \ \bm f_{c_{n_c}}, \ \dot{\bm q}_j) \in \mathbb{R}^{3 n_c + n_a}}$ the robot model can be expressed in the state space as $\dot{\bm{x}}(t) = \bm f(\bm x(t), \bm u(t), t)$.
As mentioned above, CENTAURO robot consists of 37 actuated joints which leads to a model with state dimension $n_x = 49$ and control input dimension $n_u = 55$.

%% file: Mpc.tex
\section{Whole-body Model Predictive Control}
\label{sec:mpc_formulation}
Model Predictive Control has been a popular strategy for generating robot motions through iteratively solving an OCP. Solving the OCP at each MPC iteration consists of computing the control input that minimizes an objective function over a finite-time horizon while satisfying a number of constraints (equality, inequality, dynamics). The optimized control input is applied on the robot until the OCP of the next MPC iteration is solved and a new optimal control policy is available to be used. In mathematic terms each MPC iteration can be formulated as the optimization problem in \cref{eq:OCProblem}.
\begin{equation} \label{eq:OCProblem}
    \begin{cases}
    \underset{\bm u(.)}{\min} \ \ \Phi(\bm x(T)) + \displaystyle \int_{0}^{T} l(\bm x(t), \bm u(t), t)dt \\[2ex]
    \text{s.t.} \ \ \dot{\bm{x}}(t) = \bm f(\bm x(t), \bm u(t), t) \\
    \ \ \ \ \ \bm g_1(\bm x(t), \bm u(t), t) = 0 \\
    \ \ \ \ \ \bm g_2(\bm x(t), t) = 0 \\
    \ \ \ \ \ \bm h(\bm x(t), \bm u(t), t) \geq 0 \\
    \ \ \ \ \ \bm x(0) = \bm x_0,
    \end{cases}
\end{equation}
where $\bm x(t) \in \mathbb{R}^{n_x}$ and $\bm u(t) \in \mathbb{R}^{n_u}$ are the state and control input vectors, respectively, and $\dot{\bm{x}}(t) = \bm f(\bm x(t), \bm u(t), t)$ is the dynamics model described in \cref{sec:model}. The terms $l(\bm x, \bm u, t)$ and $\Phi(\bm x(T))$ are the stage and terminal cost, respectively. The terminal cost is a common practice used to mimic the "tail" of the cost of the original infinite horizon problem. The terms $\bm g_1(\bm x(t), \bm u(t), t), \ \bm g_2(\bm x(t), t), \ \bm h(\bm x(t), \bm u(t), t), \ \bm x_0$ in \cref{eq:OCProblem} form state-input equality, state-only equality, state-input inequality and initial state constraints, respectively. In this work the Sequential Linear Quadratic (SLQ) solver, detailed in \cite{farshidian2017efficient, farshidian2017real} and available in \cite{OCS2} is used with selected time horizon of 1 sec. The SLQ algorithm is a DDP variation in the continuous-time domain. As common practise in real-time MPC iteration schemes \cite{diehl2005real}, at each MPC iteration only the solution of the first SLQ iteration is obtained.

\subsection{Constraints}
A shown in \cref{eq:OCProblem}, the MPC formulation considers the robot dynamics and a number of equality and inequality constraints that are handled through a projection and an augmented-Lagrangian method \cite{sleiman2021constraint}, respectively. These constraints are summarized in \cref{tab:constraints}. For the rest of this paper time dependency is omitted for simplicity.

\subsubsection{Contact-related constraints} Given a user desired gait (i.e. the contact state of all EEs along the time horizon), friction cone (inequality) and contact complementarity (equality) constraints are imposed for the EEs. Differently from leg EEs the velocity of the arm EEs in contact is not constrained to zero in order to include potential interaction with movable objects. It is worth mentioning that the friction cone constraint includes a regularization term which provides as well a level of stability robustness by imposing a positive lower bound at the forces along the normal direction of the contact surface. Detailed mathematical description of the contact-related constraints is omitted, since they currently consist textbook knowledge for legged planning and control, however the reader can refer to \cite{sleiman2021unified} for the details. 

\subsubsection{Swing leg EE normal velocity} For each leg EE motion tracking along the normal direction is achieved through a state-input equality constraint, shown in \cref{eq:ee_equality_constraint}. The horizontal components of the motion are not constrained.
\begin{equation}
    v_{ee}^z(\bm x, \bm u) - v_{ee}^{ref,z} + \alpha_{p} \cdot (p_{ee}^z(\bm x) - p_{ee}^{ref,z}) = 0 
\label{eq:ee_equality_constraint}
\end{equation}
where $p_{ee}^z(\bm x)$ and  $v_{ee}^z(\bm x, \bm u)$ are the corresponding normal components of the state-dependent and state-input-dependent position and velocity of the leg EE, respectively. The superscript $(.)^{ref}$ indicates the desired values at each time according to the desired trajectory and $\alpha_p$ is a tuned gain that is multiplied by the position error. Feedback on the position is necessary since a velocity-only constraint would result in drifting. A constraint of this type can be as well used for tracking the motion of the arm EEs, as well. Due to the limited space only one of the showcased arm EE motions in \cref{sec:results_main} is evaluated using such kind of constraints while for the rest the MPC formulation makes use of cost terms to track the arm EE motion, as described in \cref{sec:cost_function}.
%
\begin{table}[b]
\caption{Constraints included in the MPC formulation}
\label{tab:constraints}
\begin{center}
\begin{tabular}{|c||c|c|}
\hline
constraint & Type & Complexity\\
\hhline{|=||=|=|}
robot dynamics & state-input equality & nonlinear \\
\hline
friction cone & input inequality & nonlinear \\
\hline
zero force at free-motion & input equality & linear \\
\hline
contact leg EE zero velocity & state-input equality & nonlinear \\
\hline
swing leg EE normal velocity & state-input equality & nonlinear \\
\hline
arm joint position limits & state inequality & linear \\
\hline
\end{tabular}
\end{center}
\end{table}
\subsection{Cost function}
\label{sec:cost_function}
The cost function for the OCP is selected to be:
\begin{align}
l(\bm x, \bm u)
= \, ||\bm x - \bm x^{ref}||^2_{\bm Q} + ||\bm u - \bm u^{ref}||^2_{\bm R_j + \bm R_{ts}} + \notag
\\
+ ||\bm p_{ee}(\bm x) - \bm p_{ee}^{ref}||^2_{\bm Q_{e_P}} + ||\bm e_{O}(\bm x)||^2_{\bm Q_{e_O}} + l_{sc}(\bm x)
\label{eq:cost_function}
\end{align}
where $||\dot{ \ }||^2_H$ denotes the weighted (with weight matrix $H$) squared $L_2$ norm. The state is regularized with weight matrix $\bm Q$ while the weight matrix related with the control input consists of a positive definite sum of two terms $\bm R_j + \bm R_{ts}$. The matrix $\bm R_j \in \mathbb{R}^{n_u \times n_u}$ is a positive definite weight matrix for regularizing all the control inputs while $\bm R_{ts} \in \mathbb{R}^{n_u \times n_u}$ is related with task specifications, as described below.

\subsubsection{Leg end-effector velocity}
\begin{figure}[t]
  \centering
  \graphicspath{{figures/}}
  \includegraphics[width=0.95\columnwidth]{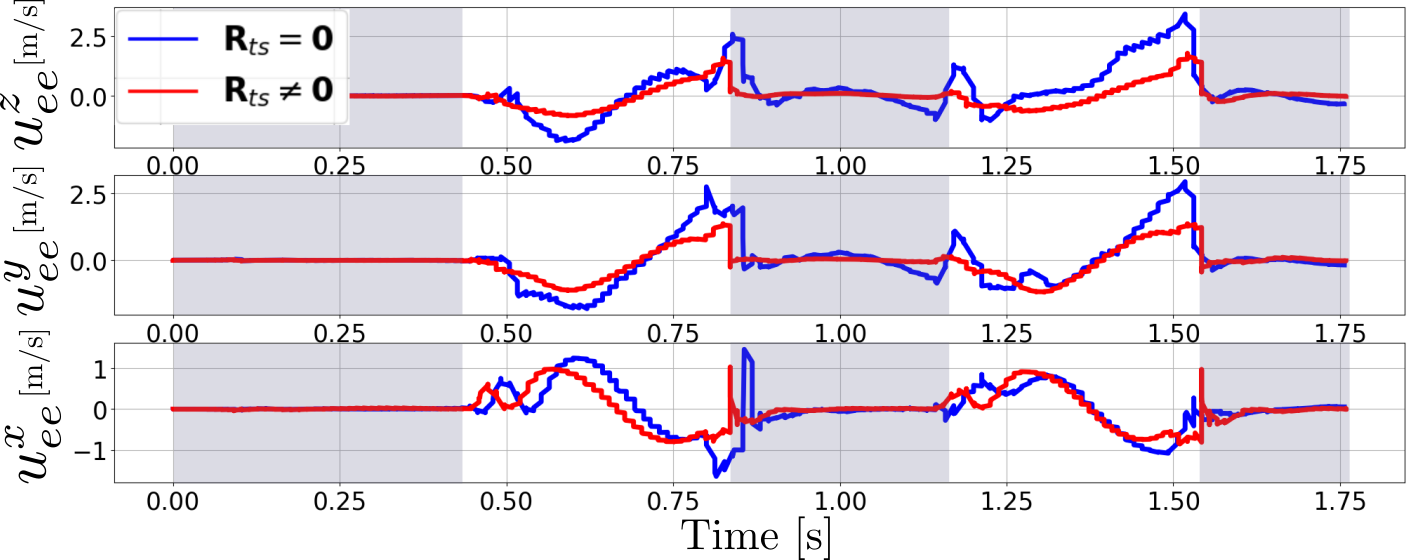}
  \caption{Cartesian velocity of a leg EE (rear right) with (red) and without (blue) the proposed cost term during a dynamic trotting motion in simulation. Penalizing cartesian leg EE velocities results in reducing impacts during leg touch-down. The shaded regions correspond to leg in contact.}
  \label{Fig:cost_ee_velocity}
\end{figure}

$\bm R_{ts}$ is selected such that the task space (cartesian) velocity of the leg EEs wrt the base link are penalized. This is achieved by mapping a weight matrix $\bm R_{leg,ee} \in \mathbb{R}^{3n_{c,leg} \times 3n_{c,leg}}$ from the cartesian space into the joint space through the stacked Jacobians of the leg EEs wrt the base $\bm J_{c,leg} \in \mathbb{R}^{3 n_{c,leg}\times (n_{c,leg} \cdot n_{a,leg})}$, as shown in \cref{eq:control_input_weight}. This input cost affects only the leg joints, hence the zero submatrices corresponding to the forces and upper body joints. The number of leg EEs is $n_{c,leg}=4$ with $n_{a,leg}=6$ actuated joints per leg. The jacobians are computed at the nominal joint configuration which is included in $\bm x^{ref}$.
\begin{align}
\bm R_{ts} = 
\begin{bmatrix}
    \bm 0 & \bm 0 & \bm 0 \\
    \bm 0 & \bm J_{c,leg}(\bm x^{ref})^T \bm R_{leg,ee} \bm J_{c,leg}(\bm x^{ref}) & \bm 0 \\
    \bm 0 & \bm 0 & \bm 0
\end{bmatrix}
\label{eq:control_input_weight}
\end{align}
The cost term related with the leg EE cartesian velocities is particularly important for stepping motions, since large velocities can create large impacts at leg touch-down. The horizontal components are as well penalized since they can produce lateral disturbance to the robot. This is demonstrated in \cref{Fig:cost_ee_velocity} with data from a dynamic trotting motion in simulation. In this simulation the robot was not able to continue stepping without the proposed cost term.
\subsubsection{Arm EE pose tracking} The weights $\bm Q_{e_P}, \ \bm Q_{e_O} \in \mathbb{R}^{3 \times 3}$ penalize the position and orientation error between the arm EE frame and the desired pose at each time instant, $\bm p_{ee}(\bm x) - \bm p_{ee}^{ref}$ and $\bm e_O(\bm x)$ respectively. The orientation error is computed as follows \cite{siciliano2008robotics}:
\begin{equation}
    \bm e_O(\bm x) = \eta_e(\bm x) \bm \epsilon_{ref} - \eta_{ref} \bm \epsilon_e(\bm x) - \bm S(\bm \epsilon_{ref}) \bm \epsilon_e(\bm x)
\end{equation}
where $\bm Q_e(\bm x) = \{\eta_e(\bm x), \bm \epsilon_e(\bm x)\}$, $\bm Q_{ref} = \{\eta_{ref}, \bm \epsilon_{ref}\}$ are unit quaternions of the achieved and desired orientation, respectively, and $\bm S(\cdot)$ the skew-symmetric matrix operator.
\subsubsection{Self-collision avoidance} The term $l_{sc}(\bm x)$ in \cref{eq:cost_function} is a state cost for self-collision avoidance which is particularly important for robots with multiple kinematic chains. The advantage of such a consideration within a MPC, in contrast to instantaneous approaches, is that collision-free motions are optimized along the receding horizon thus self-collision distances are predicted. In this work, the approach of \cite{chiu2022collisionfree} is used where the cost is the sum of relaxed barrier function evaluations, one for each pair of links to be considered. This way the cost is acting as a soft constraint; more details can be found in the corresponding publication. For CENTAURO multiple link pairs are considered, mainly between the arm links and the robot's torso, base link as well as the front leg on the same side (e.g. between left arm and left leg).
The MPC formulation considers 22 link pairs for each arm and the rest of the robot body, as well as the EE links of the two arms. This makes up a total number of 45 link pairs.

The terminal cost $\Phi(\bm x(T))$ of \cref{eq:OCProblem} includes only a term related to arm EE tracking (similar to the one described above) to make sure that the MPC does not continuously postpone tracking until after the end of the receding horizon.

%% file: FromMPCToMotorControl.tex
\section{From MPC to Motor Control}
\label{sec:low_level_modules}
This section describes how the optimal MPC state and control input policy is interfaced with the low-level joint impedance controllers of the joint actuators.
\subsection{Low-level reference generator}
\label{sec:torque_mapping}
The whole-body MPC of \cref{sec:mpc_formulation} is interfaced with the joint impedance controllers of the robot through a low-level reference generator. The reference generator is responsible for sampling the latest optimal MPC policy at 500 Hz (using linear interpolation where needed along the time horizon) and mapping the optimal state and control input vectors to joint impedance controller references for the actuated joints, namely feedforward torque $\bm \tau^{ff}_{j} \in \mathbb{R}^{n_a}$, joint position $\bm q^{ref}_j \in \mathbb{R}^{n_a}$ and joint velocity $\dot{\bm q}^{ref}_j \in \mathbb{R}^{n_a}$ references, respectively. These are then sent to the controller of \cref{sec:joint_impedance}.

The position and velocity references $\bm q^{ref}_j, \ \dot{\bm q}^{ref}_j$ can be directly accessed as part of the sampled optimal state and control input vectors, respectively. The feedforward torque vector $\bm \tau^{ff}_j$, differently from previous works that use an instantaneous WBC \cite{sleiman2021unified}, is computed by solving the unconstrained inverse dynamics. This is efficiently done using the Recursive Newton-Euler Algorithm (RNEA), an implementation of which is available in Pinocchio library \cite{carpentier2019pinocchio}. In practise, this consists a simple mapping of the MPC policies to torques without any additional feedback as in WBC. Given the generalized reference configuration $\bm q^{ref} \in \mathbb{R}^{6 + n_a}$, velocity $\dot{\bm q}^{ref} \in \mathbb{R}^{6 + n_a}$, acceleration $\ddot{\bm q}^{ref} \in \mathbb{R}^{6 + n_a}$ and external forces $\bm f_c \in \mathbb{R}^{3n_c}$ the RNEA computes the actuation torques $\bm \tau^{ff}_j$ that satisfy \cref{eq:inverse_dynamics}. $\bm M(\bm q) \in \mathbb{R}^{(6 + n_a)\times(6 + n_a)}$ is the inertia matrix, $\bm h(\bm q^{ref}, \dot{\bm q}^{ref}) \in \mathbb{R}^{6 + n_a}$ is the vector of centrifugal, Coriolis and gravity forces, $\bm S_a \in \mathbb{R}^{n_a \times (6 + n_a)}$ the actuated joint selection matrix and $\bm J_c \in \mathbb{R}^{3n_c\times(6 + n_a)}$ includes the Jacobians mapping external forces to generalized coordinates. The generalized acceleration vector is not available at the MPC and joint acceleration is not measurable on the actuators, thus zero generalized accelerations $\ddot{\bm q}^{ref} = \bm 0$ are assumed in \cref{eq:inverse_dynamics}. This assumption is compensated by the feedback part of the control law in \cref{eq:joint_impedance}.
\begin{align}
    \bm S_a^T \bm \tau^{ff}_j &= \bm M(\bm q^{ref})\ddot{\bm q}^{ref} + \bm h(\bm q^{ref}, \dot{\bm q}^{ref}) - \bm J_c^T \bm f_c \label{eq:inverse_dynamics}
\end{align}  
\subsection{Joint Impedance Control}
\label{sec:joint_impedance}
In contrast to previous optimization-based work on CENTAURO \cite{dadiotis2022trajectory}, the actuated joints are controlled in joint impedance mode, described by \cref{eq:joint_impedance} in a vector form.
\begin{align}
    \bm \tau^{ref}_j &= \bm \tau^{ff}_j + \bm K_p \cdot (\bm q^{ref}_j - \bm q_j^m) + \bm K_d \cdot (\dot{\bm q}^{ref}_j - \dot{\bm q}_j^m)
    \label{eq:joint_impedance}
\end{align}
where $\bm q_j^m, \ \dot{\bm q}_j^m \in \mathbb{R}^{n_a}$ consist the vectors of measured joint positions and velocities. These controllers are running at 2 kHz. The feedback gain matrices $\bm K_p, \ \bm K_d \in \mathbb{R}^{n_a \times n_a}$ express stiffness and damping, respectively, at the joint level. Stiffness was tuned by trial and error to be $30, \ 120, \ 200 \ N m/rad$ and damping $30, \ 60, \ 60 \ N m \cdot s/rad$ for the small, medium and large-size motors of CENTAURO, respectively. Stiffness is set at relatively low levels since the reference joint positions and velocities have been already considered for the generation of the feedforward torque in \cref{eq:inverse_dynamics}. For what concerns the velocity controlled wheel joints, only velocity references $\bm \dot{q}^{ref}_j$ are passed to them which are tracked using a proportional feedback at the velocity level. Thus, the control law of each wheel motor can be still described by a scalar version of \cref{eq:joint_impedance} with $\tau^{ff}_j = 0 \ N m, \ K_p = 0 \ N m/rad$ with selected damping $K_d = 10 \ N m \cdot s/rad$.

\subsection{State Estimation}
\label{sec:state_estimation}
The state estimation module consists of an algorithm which estimates the robot base pose and twist using IMU, joint position and velocity measurements. Upon estimating the base pose and twist, an estimate of the OCP state $\hat{\bm x} \in \mathbb{R}^{n_x}$ is computed which includes the centroidal momentum. The computed state estimate is then passed to the whole-body MPC to be used as an initial condition in case a new MPC iteration is about to start. The state estimator needs to be aware of the contact state of the EEs and so the user-selected gait command sent to the whole-body MPC is as well passed to the state estimator module.

%% file: Results.tex
\section{Experiments \& Evaluation}
\label{sec:results_main}
\subsection{Implementation}
\label{subsec:implementation}
The MPC is developed using the OCS2 toolbox \cite{OCS2} which provides an implementation of the used SLQ algorithm and automatic differentiation capabilities based on CppADCodeGen for the derivatives of the dynamics, cost and constraints. The toolbox relies on Pinocchio \cite{carpentier2019pinocchio} for the dynamics and kinematics. Synchronization mechanisms regarding the different processes that an MPC-based pipeline requires (e.g. running the solver, policy sampling, observation update) are provided. To enable deploying with Gazebo simulator and the real platform the MPC is interfaced through ROS with the XBot2 middleware \cite{LAURENZI2023104379} used for controlling CENTAURO.
\subsection{Experiments}
\label{sec:experiments}
In this section the conducted experiments are described, which can be found in the accompanying video. These are evaluated in \cref{sec:evaluation}. The OCP problem of each experiment is different due to the different constraints/costs of the executed task.
\subsubsection{Arm end-effector free motion }
\label{subsec:free_motion}
The performance of the framework is firstly evaluated with a whole-body motion where the left arm EE of the robot is tracking a cartesian linear motion while all leg EEs are in contact with the ground. Position tracking for the arm EE is achieved using using two different approaches: the cost term in \cref{eq:cost_function} and equality constraints similar to \cref{eq:ee_equality_constraint}. The cost term related with EE orientation tracking is not included. The resulted motion highlights the advantage of whole-body coordination for increasing the robot's workspace since the robot reaches poses close to the ground and up to 2 m height (see \cref{Fig:robot_photo}b).
\subsubsection{Heavy object retrieval from ground}
\label{subsec:object_ground}
\begin{figure}[b]
  \centering
  \graphicspath{{figures/}}
  \includegraphics[width=0.99\columnwidth]{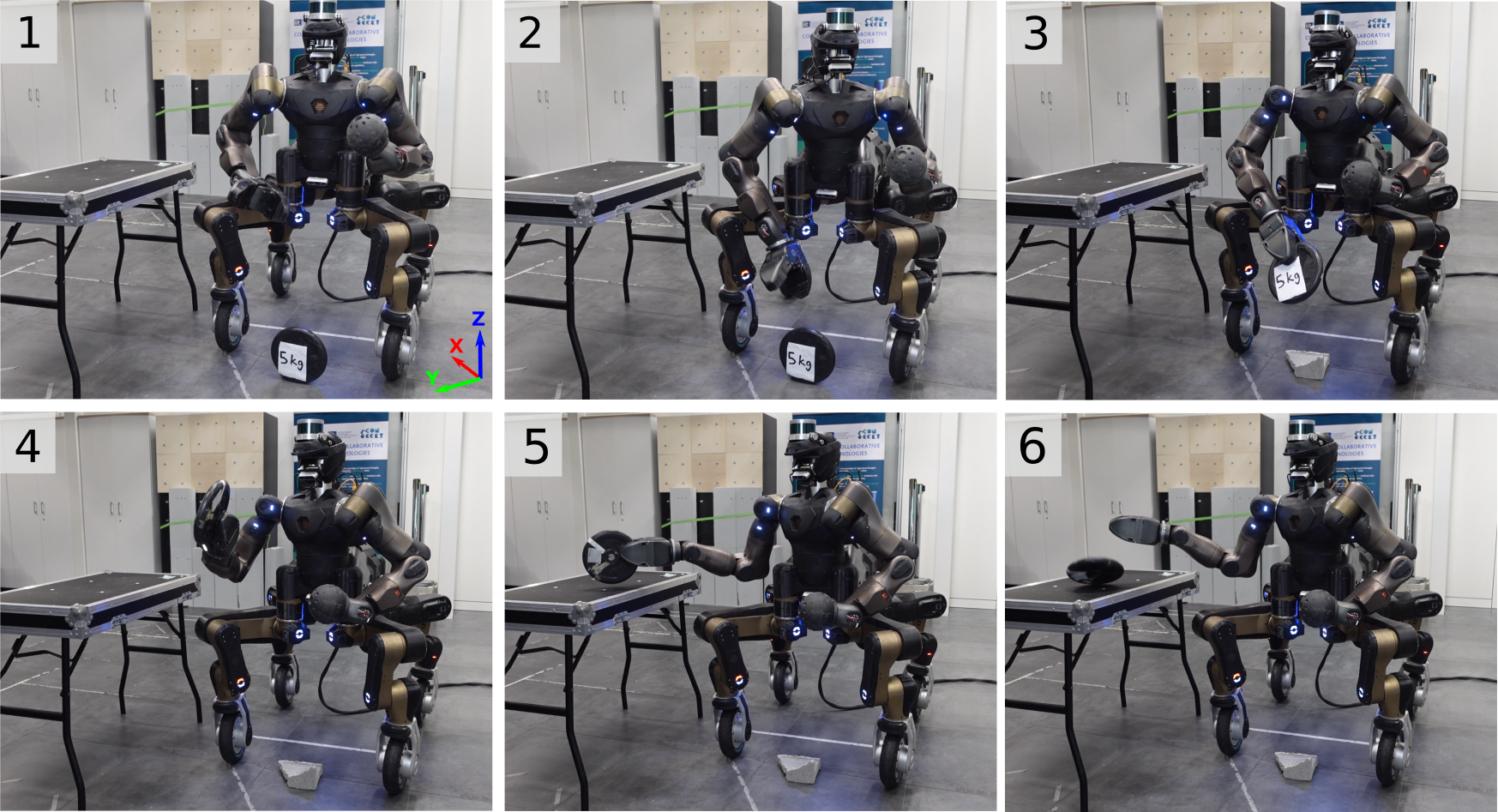}
  \caption{Snapshots of the heavy object retrieval from ground experiment described in \cref{subsec:object_ground}}
  \label{Fig:heavy_object_shots}
\end{figure}
The second experiment involves a manipulation task where the robot reaches a 5 kg object on the ground with its arm EE, grasps it (using a suitable gripper) and places it on top of a table, as shown in \cref{Fig:heavy_object_shots}. This task is challenging since the object has a significant mass and, thus, the force exerted from the object to the arm EE has to be considered in the MPC. Moreover, both arm EE position and orientation tracking costs are included in \cref{eq:cost_function}. In terms of tuning, position tracking errors are assigned with a slightly higher cost wrt orientation. The used gripper has a rather large aperture and, thus, can achieve grasping of the object despite any small angular misalignment while this way the MPC can take advantage of a less stiff orientation tracking for avoiding self-collision when needed. For EE tracking equality constraints similar to \cref{eq:ee_equality_constraint} can be used instead of the cost terms, however this is omitted due to lack of space. The video demonstrates how the arm fastly moves towards the object and close to the robot legs without self-colliding. This natural behavior of the MPC emerges from the corresponding self-collision related state cost term while the user only sends a desired trajectory (linear interpolation between the current and target pose) without worrying for potential self-collisions.
\subsubsection{Dynamic stepping}
\label{subsec:dynamic_locomotion}
A number of dynamic stepping experiments are also presented where the robot is performing crawling and trotting without including any arm EE motion tracking. Self-collision avoidance and arm joint position limits are, also, not included to simplify the OCP. The arms remain always close to the reference configuration due to the state regularization cost. The stepping motions mainly include leg swings of 0.3 sec duration. For a CENTAURO robot, where leg distal mass is significant, this is quite challenging and is achieved for first time on the real hardware.
\subsubsection{Loco-manipulation in simulation}
\label{subsec:locomanipulation}
A highly demanding motion where the robot is dynamically trotting and at the same time following a reference position trajectory with its arm EE is presented in a full-physics numerical simulator (Gazebo). This motion uses the centroidal dynamics formulation with self-collision and position tracking cost terms. Such an experiment is not presented on the real hardware but consists a straightforward extension of this work towards combining legged mobility with whole-body manipulation.
\subsection{Evaluation}
\label{sec:evaluation}
\begin{table}[t]
\caption{MPC achieved frequency (Hz) in simulation - same dynamics model used for both planning and as rollout instance \\(SRBD / Centroidal Dynamics)}
\label{tab:frequency_rviz}
\begin{center}
  \begin{tabular}{|p{1.8cm}||p{1.5cm}|p{1.5cm}|p{0.7cm}|p{0.65cm}|}
  \hline
   & \vspace{0.02cm} arm EE free motion with position tracking cost & arm EE free motion with position tracking constraint & \vspace{0.48cm} crawl & \vspace{0.5cm} trot\\
   \hhline{|=||=|=|=|=|}
    full formulation            & 38.3 / 28.3 & 48.5 / 27.7 &                             &             \\ \cline{1-3}
    $l_{sc} = 0$                & 44.9 / 35.0 & 58.0 / 34.3 &                             &             \\ \cline{1-3}
    w/o arm joint position limits   & \vfill 40.8 / 31.8 & \vfill 51.0 / 30.3 & \vspace{-0.5cm} 34.9 / 21.8 & \vspace{-0.5cm} 25.9 / 17.3 \\ \hline
   \end{tabular}
\end{center}
\end{table}
\begin{table}
\caption{MPC achieved frequency (Hz) on the real robot}
\label{tab:frequency_hw}
\begin{center}
    \begin{tabular}{|p{1.5cm}|p{1.5cm}|p{0.5cm}|p{0.5cm}|p{1.5cm}|}
    \hline
     arm EE free motion with position tracking constraint & \vspace{0.01cm} arm EE free motion with position tracking cost & \vspace{0.48cm} crawl & \vspace{0.5cm} trot & \vspace{0.3cm} heavy object retrieval\\
    \hhline{|=|=|=|=|=|}
    \hspace{0.4cm} 10.0 & \hspace{0.4cm} 22.4 & 18.0 & 44.0 & \hspace{0.4cm} 10.0\\
    \hline
    \end{tabular}
\end{center}
\end{table}
\subsubsection{Computational efficiency}
\label{subsec:computational_efficiency}
The MPC is first evaluated in simulation on a machine with an Intel Core i7-10700 CPU @ 2.90GHz using the same dynamics for both planning and rollout. The achieved MPC frequencies (mean values) are shown in \cref{tab:frequency_rviz}. Simulations of the arm EE free motion experiment are performed using two different approaches for tracking the arm EE position: through the position cost in \cref{eq:cost_function} and three equality constraints similar to \cref{eq:ee_equality_constraint}. For both cases the SRBD and the centroidal dynamics models as well as the computational overhead of including self-collision avoidance and joint position limits are evaluated. Dynamic stepping motions are, as well, included in the table. As expected, including additional costs/constraints and/or more complex dynamics in the formulation decelerates the SLQ solver which, however, manages to achieve frequencies more than 17 Hz in all cases. Difference in the achieved performance (especially for the SRBD model) can be noticed between the tracking cost and constraint cases which may be, also, explained by the fact that different tunings were needed to render the motion successful.

Next, the computational efficiency with the real robot in the loop is evaluated in real experiments. The focus in on using centroidal dynamics since it is a higher fidelity model that considers the updated inertia on each robot configuration as well as the momentum due to the joint velocities. The achieved frequencies on CENTAURO's Intel Core i9-10900K CPU @ 3.70 GHz are shown in \cref{tab:frequency_hw}. As seen, the whole-body MPC achieves replanning at 10 Hz in the most complex scenario of heavy object retrieval (where arm EE position and orientation is controlled). For other motions this frequency can increase up to 22.4 Hz. It is noted that from all the included motions, trotting is the only one for which the MPC considers the SRBD and not the centroidal dynamics. By the time of writing this paper these have been the best results we have managed to record. 
\subsubsection{Low-level reference generation \& joint impedance control}
\label{subsec:joint_impedance_evaluation}
\begin{figure}[b]
  \centering
  \graphicspath{{figures/}}
  \includegraphics[width=0.8\columnwidth]{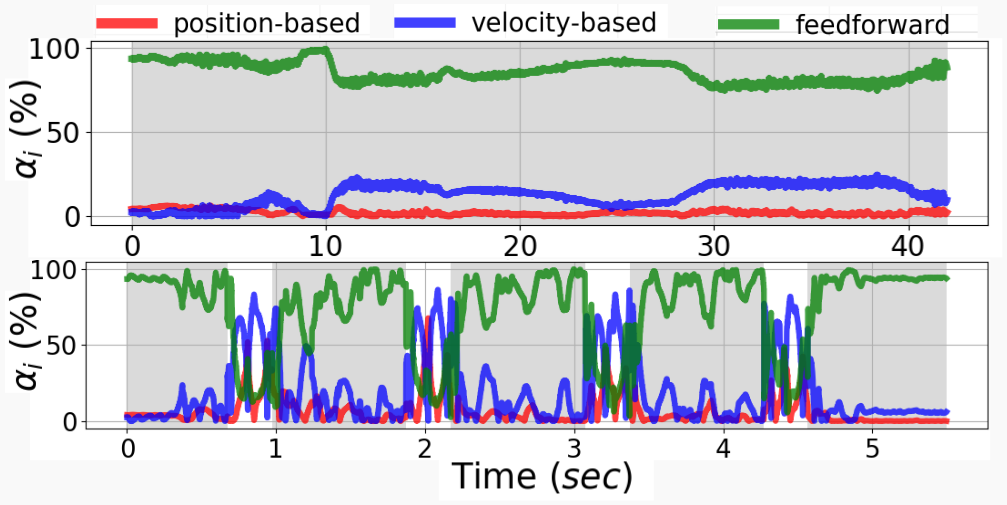}
  \caption{Contribution of the different torque terms (position-based feedback, velocity based feedback and feedforward torque) to the generated torque reference during free motion experiment (top) and the dynamic crawl stepping (bottom). The shaded regions denote leg in contact. Only data from the front left hip pitch joint are presented.}
  \label{Fig:torque_percentage}
\end{figure}

The low-level reference generator and joint impedance controller consist an important part of the pipeline rendering the MPC optimal policy useful for the torque controlled joints. To understand their behavior, for each joint the modulus of each torque term in \cref{eq:joint_impedance} (i.e. torque term based on position feedback, velocity feedack and torque feedforward, respectively) is calculated as a percentage $\alpha_i$ of the sum of the modulus of all terms, as shown in \cref{eq:torque_percentage}. 
\begin{align}
    \alpha_i = \frac{100 \ \% \cdot |i|}{\sum_{k \in S} |k|}
    \label{eq:torque_percentage}
\end{align}
where $i \in S = \{\tau_j^{ff}, \ K_p \cdot (q_j^{ref} - q_j^m), \ K_d \cdot (\dot{q}_j^{ref} - \dot{q}_j^m)\}$. In \cref{Fig:torque_percentage} data from a single joint (the front left leg hip pitch joint) of the robot are included. The top part of the \cref{Fig:torque_percentage} shows that during an arm EE motion the feedforward torque originated from the RNEA algorithm contributes most on the total reference that is the output of the joint impedance control law. The velocity-based torque term has a significant contribution at some parts of the motion where the joints need to accelerate significantly. Since generalized accelerations are neglected at the low-level reference generator, the feedback terms, and mainly the velocity-related term, of the joint impedance control are compensating for them whenever high acceleration has to be achieved. The bottom part of the figure demonstrates that during a dynamic stepping motion the feedback part becomes significantly important. In particular, during the swing phases of the leg, the feedforward contribution is diminished and most of the torque reference is due to the feedback part. This is explained due to the need for achieving high acceleration on the swinging leg joints. Another reason for the significant contribution of the velocity-related term is due to the velocity jumps that characterize the lift off and touch down events.
\subsubsection{Joint level tracking}
\label{subsec:joint_level_tracking}
\begin{figure}
  \centering
  \graphicspath{{figures/}}
  \includegraphics[width=\columnwidth]{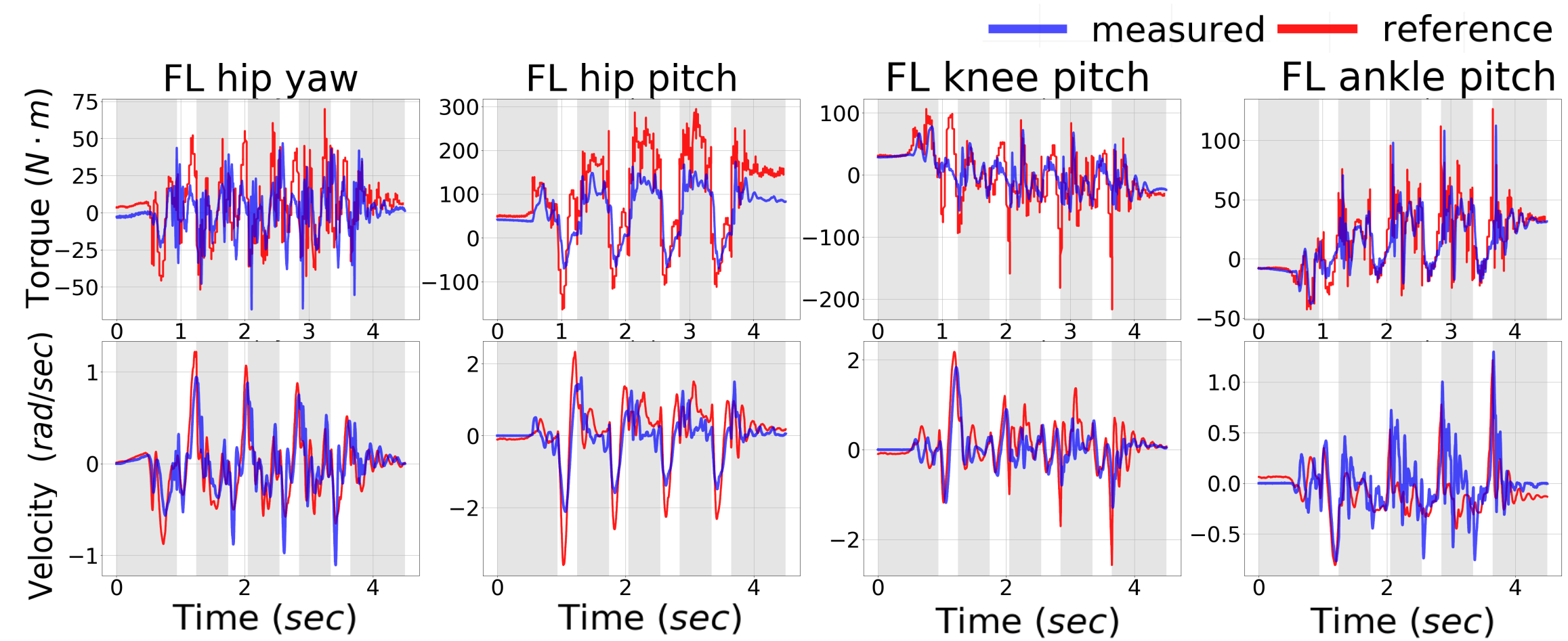}
  \caption{Joint torque and velocity tracking for the front left (FL) leg joints during a trotting motion. The ankle yaw and wheel joints are omitted since they exhibit minimal motions during stepping, as can be easily understood by the leg kinematics. Shaded regions correspond to leg in contact.}
  \label{Fig:stepping_experiment}
\end{figure}

Joint torque and velocity tracking of a single leg during the trotting motion, which consists the most dynamic motion achieved on the hardware, is shown in \cref{Fig:stepping_experiment}. Joint velocities are part of the MPC control input and, in a sense, have to be tracked adequately. The torque tracking plots demonstrate how the overall framework explores the large torque capacity of the actuators when a dynamic motion requires it, especially at the pitch joints since these are highly contributing to leg lifting and landing. Leveraging this range of joint torque capacity for dynamic stepping has not been possible using the framework of previous work \cite{dadiotis2022trajectory} since, in practice, the robot was losing stability at the very first dynamic steps. Although torque limits are not explicitly satisfied at the MPC, it is worth mentioning that during the presented motions all references are within the actuation limits. Despite this, there are significant torque tracking errors at the hip pitch joint, probably due to limited torque controller bandwidth at high torque amplitudes. 
\subsubsection{Arm EE motion tracking}
\label{subsec:arm_ee_tracking}
\begin{figure}
  \centering
  \graphicspath{{figures/}}
  \includegraphics[width=0.9\columnwidth]{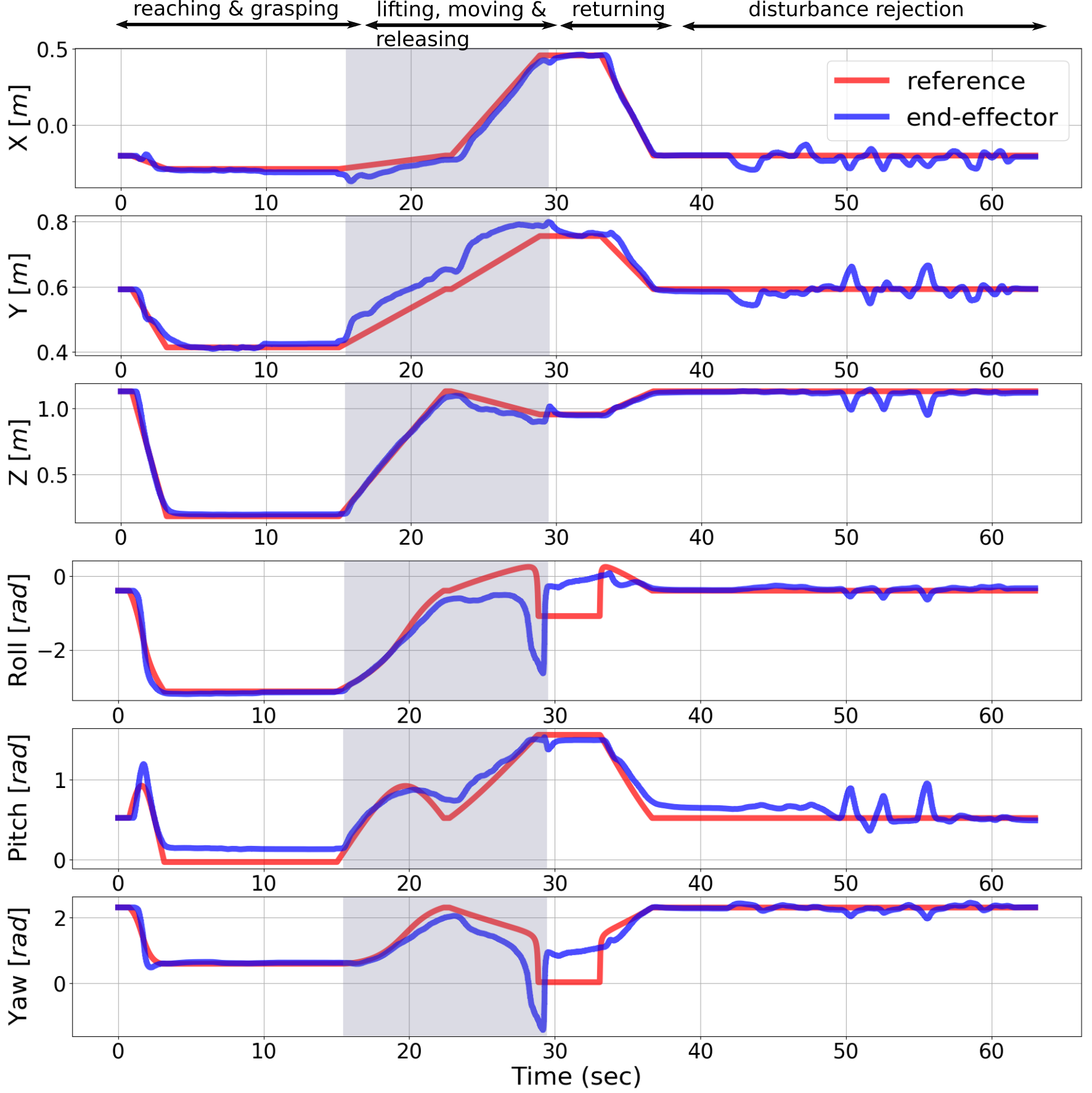}
  \caption{Reference (red) and achieved (blue) arm EE pose during the heavy object retrieval task. The shaded region denotes arm in contact with the object. The reference frame is shown in \cref{Fig:heavy_object_shots}.}
  \label{Fig:ee_tracking}
\end{figure}
In \cref{Fig:ee_tracking} the arm EE motion tracking during the heavy object retrieval experiment is shown. The references sent to the MPC (red color) consist a linear interpolation and spherical linear interpolation (for the position and quaternion orientation, respectively) between discrete target poses defined for this task. Notice that orientation in \cref{Fig:ee_tracking} is depicted in roll-pitch-yaw angles where any rapid variation does not imply non-smooth reference. As specified in \cref{sec:mpc_formulation}, the MPC formulation was on purpose designed for slightly stricter position wrt orientation tracking, which results in larger steady state errors in orientation. Another interesting observation is that the tracking capability degrades while the robot is carrying the 5 kg object. This is, partially, expected since the model of the object is unknown. Succeeding the task only relies on the limited ability of the MPC to predict the external force exerted at the arm EE through continuously receiving state observations and replanning. However, there are significant errors in the X and Y directions and orientations since the MPC compromises tracking in this directions to avoid self-collision with the leg. Finally, there is an impact at the time of releasing the object which causes tracking errors. At the last part, the user disturbs the robot body and arm EE, as can be seen in the accompanying video. Since low stiffness gains were used at the joint impedance controller the robot behaves relatively compliant. Most importantly the plot demonstrates that the achieved MPC frequency, which has an order magnitude of tens of Hz, provides the necessary feedback for achieving EE motion tracking and constraint satisfaction.

%% file: Conclusion.tex
\section{Conclusion and Future Work}
This work presents the experimental results of controlling the highly redundant dual-arm quadruped CENTAURO with whole-body MPC. This is the highest DoF robot that has been controlled in real-time with whole-body MPC so far. At the lower-level, the MPC plans are interfaced with the joint impedance controllers without the need of a WBC. The results indicate that the approach can be successfully applied to control the motion of highly complex robots which are, traditionally, teleoperated or controlled with simplified techniques. Real-time performance and replanning frequency of tens of Hz are achieved for a variety of complex tasks i.e. dynamic stepping and heavy object manipulation.

For future work, deploying the Riccati-like feedback gains provided by the solver could potentially increase the state feedback frequency \cite{dantec2022first ,mastalli2022agile, grandia2019feedback}. In terms of targeted tasks, using MPC for manipulation of unknown environments remains an open challenge, which few works have tried to address \cite{minniti2021model, arcari2023bayesian}. Another direction is to investigate how joint torque sensing can be considered within the proposed MPC where joint torques are not explicitly available.

%% file: references.tex
\bibliographystyle{IEEEtran}
\bibliography{bibliography}